%% file: fila2020.tex
\documentclass[conference]{IEEEtran}
\usepackage{cite}
\usepackage{amsmath,amssymb,amsfonts}
\usepackage{algorithmic}
\usepackage{graphicx}
\usepackage{textcomp}
\usepackage{xcolor}
\usepackage{booktabs}
\usepackage{multirow}
\usepackage{comment}

\usepackage{hyperref}
\hypersetup{
    urlcolor  = black,
	colorlinks = true,
	citecolor = blue, 
	linkcolor = blue 
}

\newcommand{\ra}[1]{\renewcommand{\arraystretch}{#1}}

\begin{document}

\title{BeFair: Addressing Fairness in the Banking Sector}


\author{\IEEEauthorblockN{Alessandro Castelnovo\IEEEauthorrefmark{1},
Riccardo Crupi\IEEEauthorrefmark{1},
Giulia Del Gamba\IEEEauthorrefmark{3},
Greta Greco\IEEEauthorrefmark{1},\\
Aisha Naseer\IEEEauthorrefmark{2},
Daniele Regoli\IEEEauthorrefmark{1},
Beatriz San Miguel Gonzalez\IEEEauthorrefmark{2}}
\IEEEauthorblockA{\IEEEauthorrefmark{1}\textit{Data Science and Artificial Intelligence, Intesa Sanpaolo}}
\IEEEauthorblockA{\IEEEauthorrefmark{3}\textit{European Regulatory and Public Affairs, Intesa Sanpaolo}\\
Turin, Italy\\
Email: name.surname@intesasanpaolo.com}
\IEEEauthorblockA{\IEEEauthorrefmark{2}\textit{Fujitsu Laboratories of Europe}\\
London, UK\\
Email: name.surname@uk.fujitsu.com}
}

\maketitle

\begin{abstract}

Algorithmic bias mitigation has been one of the most difficult conundrums for the data science community and Machine Learning (ML) experts. Over several years, there have appeared enormous efforts in the field of fairness in ML. Despite the progress toward identifying biases and designing fair algorithms, translating them into the industry remains a major challenge. In this paper, we present the initial results of an industrial open innovation project in the banking sector: we propose a general roadmap for fairness in ML and the implementation of a toolkit called \texttt{BeFair} that helps to identify and mitigate bias. Results show that training a model without explicit constraints may lead to bias exacerbation in the predictions.\footnote{
© 2021 IEEE.  Personal use of this material is permitted.  Permission from IEEE must be obtained for all other uses, in any current or future media, including reprinting/republishing this material for advertising or promotional purposes, creating new collective works, for resale or redistribution to servers or lists, or reuse of any copyrighted component of this work in other works.}

\end{abstract}


\section{Introduction}

The notion of fairness is still ambiguous and not uniquely defined, mainly because it is context-dependent with complex interdependencies among several attributes. The disparate nature of algorithmic bias and unfair discrimination suggests that fairness could neither be automated \cite{wachter2020fairness} nor it can be monolithic. Various Machine Learning (ML) techniques commonly exercise intuitively unfair behaviours, typically due to amplification of bias already encoded in the data or due to minimizing average error to fit majority populations \cite{chouldechova2018frontiers, chouldechova2020snapshot}. Anecdotal evidences suggest that a number of fairness metrics are being used to assess the inherent bias in data. Generally, fairness is treated at two different levels: group fairness and individual fairness~\cite{hutchinson201950}, the first trying to protect vulnerable groups of people and the second focusing on the equality of treatment at the individual level. Nevertheless, it is still not clear whether these two notions are mutually compatible \cite{dwork2012fairness, binns2020apparent}. The choice of which fairness level to use depends on the context of the use case or the concrete business problem at hand. However, most commonly fairness is applied at the group level in most domains.

Although research on fairness in ML has grown in both importance and volume over the past few years and, despite multiple metrics, approaches and methods to pursue fairness have been proposed, there is a lack of consensus on normative standards and industrial frameworks that can enable industry professionals (including data scientists and domain experts) to mitigate bias in ML models. Hence, it becomes imperative to anticipate potential sectoral requirements on fairness in Artificial Intelligence (AI) from a human-centric perspective.

Currently, policy-makers, companies and academic institutions are making efforts towards establishing guidelines and recommendations for Ethics in AI, which includes fairness in ML. One such initiative is  AI4People\footnote{AI4People, \url{https://www.eismd.eu/ai4people/}}, a multi-stakeholder forum in Europe with global activities around the promotion of a ``good AI society''. In this context, being AI4People members, Fujitsu Laboratories of Europe and Intesa Sanpaolo are collaborating on an open innovation project that reflects a proactive participation in the development and creation of a generic roadmap for Trustworthy AI.


In this paper, we present the initial results that are obtained from our collaboration, in particular, how to implement fairness in a specific banking use case of credit lending. We propose a generic roadmap for fairness in ML that can be applied to various banking use cases. Moreover, we present a toolkit called \texttt{BeFair} (Banking, explainability and Fairness) that implements multiple fairness mitigation strategies and demonstrates comparisons among them using different metrics for a specific use case.


The paper is organised as follows. Section~\ref{sec:background} covers the existing background related to fairness metrics and mitigation techniques. Section~\ref{sec:roadmap} introduces our proposed generic roadmap to fairness in ML. Section~\ref{sec:befair} elaborates our \texttt{BeFair} toolkit to monitor and assess bias and its mitigation presenting the description of findings in credit lending use case. Section~\ref{sec:discussion} discusses our findings and results and finally section~\ref{sec:conclusion} concludes the paper.

\section{Background
\label{sec:background}}

The global prevalence of AI and its increasingly ubiquitous role in society and business, impacting on every aspect of our lives, has ushered an era where \emph{trust} has never been so important before. Owing to the growing pervasiveness and adoption of AI systems and ML models, the need for achieving Trustworthy systems has become imperative.

Consequently, multiple initiatives from policy-makers, industries and academic institutions have established ethical principles, requirements and recommendations for AI systems where fairness is one of the fundamental principles (we refer to \cite{fjeld2020principled, jobin2019global} for a complete summary of these initiatives).

Diversity, non-discrimination and fairness are considered key requirements for AI systems according to these initiatives. The aim of the non-discrimination principle is to allow all individuals an equal and fair prospect to access opportunities available in a society. Individuals who are in similar situations should receive similar treatment and not be treated less favourably simply because of a particular ``protected'' characteristic\footnote{In this paper we shall use the terms ``protected'' and ``sensitive'' inter-changeably to indicate attributes to be taken into account when dealing with fairness and discrimination issues.} that they possess (e.g. sex, sexual orientation, disability, age, race, ethnic origin, national origin and religion or belief). Indirect discrimination is present when certain characteristic or factor occurs more frequently in the population groups against whom it is unlawful to discriminate. Since algorithmic decision-making systems may be based on correlations, there is a risk to perpetuate or exacerbate indirect discrimination through stereotyping, when differential treatment cannot be justified \cite{alg_human_rights}. Financial data is prone to bias and imbalance \cite{zhang2019fairness} and a multitude of research conducted on specific AI use cases (e.g. credit loan screening applications) shows that putting into practice fairness principles in industrial processes is an open issue.

From a technological perspective, the research on fairness has been approached from two key dimensions: fairness definition (what is unfair discrimination) and bias mitigation (how unfair discrimination is reduced). Next, we describe the most relevant aspects related to each dimension.

\subsection{Fairness definition}

Habitually, fairness definitions are divided into two main categories depending on the purpose that is considered: individual and group fairness~\cite{hutchinson201950}.

Individual fairness is embodied under the principle of ``similar individuals should be treated similarly''~\cite{dwork2012fairness}. Thus, this notion focuses on comparison of individuals and it consists in ensuring that any two individuals who are similar receive equal or similar outcomes. On the other hand, group fairness is focused on requiring that people belonging to protected groups receive on average the same treatment as the whole population, and are usually expressed as the equality of some statistical measure across groups~\cite{verma2018fairness}. Therefore, group fairness aims at providing equality of treatment for groups instead of specific individuals.

To assess fairness of a ML model, a precise definition is needed. In this sense, fairness has been mathematically formalised in multiple forms and there is not a clear agreement on which definition to apply in each situation. Moreover, some of the proposed definitions are mutually incompatible and exclusive under some conditions, while some of them are related in non-trivial ways~\cite{kleinberg2016inherent, chouldechova2017fair}.

Afterwards, we describe the most common definitions of fairness with respect to groups identified by some protected attribute(s) (refer to \cite{verma2018fairness, mitchell2018prediction, hutchinson201950} for more details):

\begin{itemize}
    \item \emph{Statistical Parity or Demographic Parity (DP)} is achieved when groups have the same probability of being assigned to the positive predicted class, i.e. when the decision is independent of the sensitive feature value.

    \item \emph{Conditional Demographic Parity (CDP)} modifies DP by requiring the parity of outcomes to hold not unconditionally, but within groups given by the level of other variables (e.g. credit risk level).
 
    \item \emph{Predictive Parity (PP), Equal Opportunity (EOpp) and Equality of Odds (EO)} not only consider the prediction of the model, but also the ground truth target. In particular, PP measures the precision or probability of a positive prediction to actually be in the positive class (positive predictive value) across groups. EOpp considers the probability of a subject in a positive class to have a positive prediction (true positive rate or recall). Finally, EO requires both true positive rate and false positive rate to be equal across groups. 
\end{itemize}

The aforementioned metrics are based on statistics of observational data coming from the joint distribution of the protected ground(s), input features and target labels. Besides these statistical notions of fairness, there are also proposals of metrics focusing on the use of causal relations among variables, exploiting, together with observational information, domain and expert knowledge. See e.g. \cite{kusner2017counterfactual, kilbertus2017avoiding}.

\subsection{Bias Mitigation}

Bias mitigation refers to the process of addressing specific aspects of the ML pipeline in order to remove the effect of unfair bias. 
There are many techniques within the literature that can be roughly classified into the following categories~\cite{mehrabi2019survey, oneto2020fairness}:
\begin{itemize}
    \item \emph{Pre-processing methods} are based on the idea of removing potential unfair biases directly from the training dataset. Then, a “standard” classifier is learned on this cleaned dataset. 
    In order to “clean the dataset” there are two possible families of methods.\\
    The first consists in performing a transformation of the the feature space such that the protected information is removed, while at the same time trying to preserve as much information as possible in order to efficiently estimate the target~\cite{zemel2013learning, louizos2015variational, calmon2017optimized, kamiran2012data}. A very simple form of pre-processing is the straightforward suppression of the protected attribute(s) from the dataset: this is sometimes called Fairness Through Unawareness (FTU) and embodies a very intuitive notion of individual fairness, namely that two individuals identical in all features but the protected one(s) should be given the same decision~\cite{verma2018fairness}.
    
    The second family of methods consists in transforming the dataset by working on observations, namely relabelling or resampling some of them in order to reach group fairness~\cite{kamiran2012data}.

    \item \emph{In-processing approach} consists in enforcing a model to produce fair outcomes by adding constraints or penalties to the optimization problem, thus imposing fairness at training time. This methods are highly tailored on specific underlying models, thus difficult to generalise. See e.g. \cite{zhang2018mitigating, agarwal2018reductions}.

    \item \emph{Post-processing strategies} are focused on mitigating unfair outcomes of an already trained ML model. The basic idea is to define a new classifier as a function of the (biased) outcomes of the unmitigated model, optimizing some cost function over false positives and false negatives subject to some fairness constraint. The main reference of this type of procedure is \cite{hardt2016equality}, while other approaches can be found in \cite{lohia2019bias, pleiss2017fairness}.

\end{itemize}

Furthermore, there are mitigation strategies based on causality concepts.
The majority of the existing work on these solutions has been focused on individual fairness and can be loosely considered as pre-processing methods, since they consist in training a ML model on a transformed dataset. For example, \cite{kusner2017counterfactual} proposes a method to produce counterfactually fair outcomes, where all protected information causally impacting the decision is removed from the dataset (i.e. an individual is given the same decision that she/he would have been given in the counterfactual world where sensitive features are different). \cite{chiappa2019path} extends this counterfactual fairness, accounting for the fact that not all the causal impact of the sensitive information on the decision is in general unfair, thus mitigating only with respect to variables that are part of unfair causal paths. Another method inspired by causal reasoning but not involving counterfactuals can be found in \cite{kilbertus2017avoiding}.

Various companies and public institutions have made an effort to encompass fairness metrics and mitigation techniques through specific software tools, toolkits and checklists, such as IBM AI Fairness 360 \cite{bellamy2018ai}, Google What-If Tool \cite{wexler2019if}, Aequitas \cite{saleiro2018aequitas}, and the research to co-design AI fairness checklists \cite{madaio2020co}. Despite the progress made, these solutions are usually context-agnostic. However, each industry and process have their particularities and it is needed to research and formalize ad-hoc solutions from institutions. Next sections describe the results of our research in the banking sector and in particular for a credit-lending use case.

\section{Roadmap to Fairness\label{sec:roadmap}}


\begin{figure*}[htbp]
    \centering
    \includegraphics[width=0.8\textwidth]{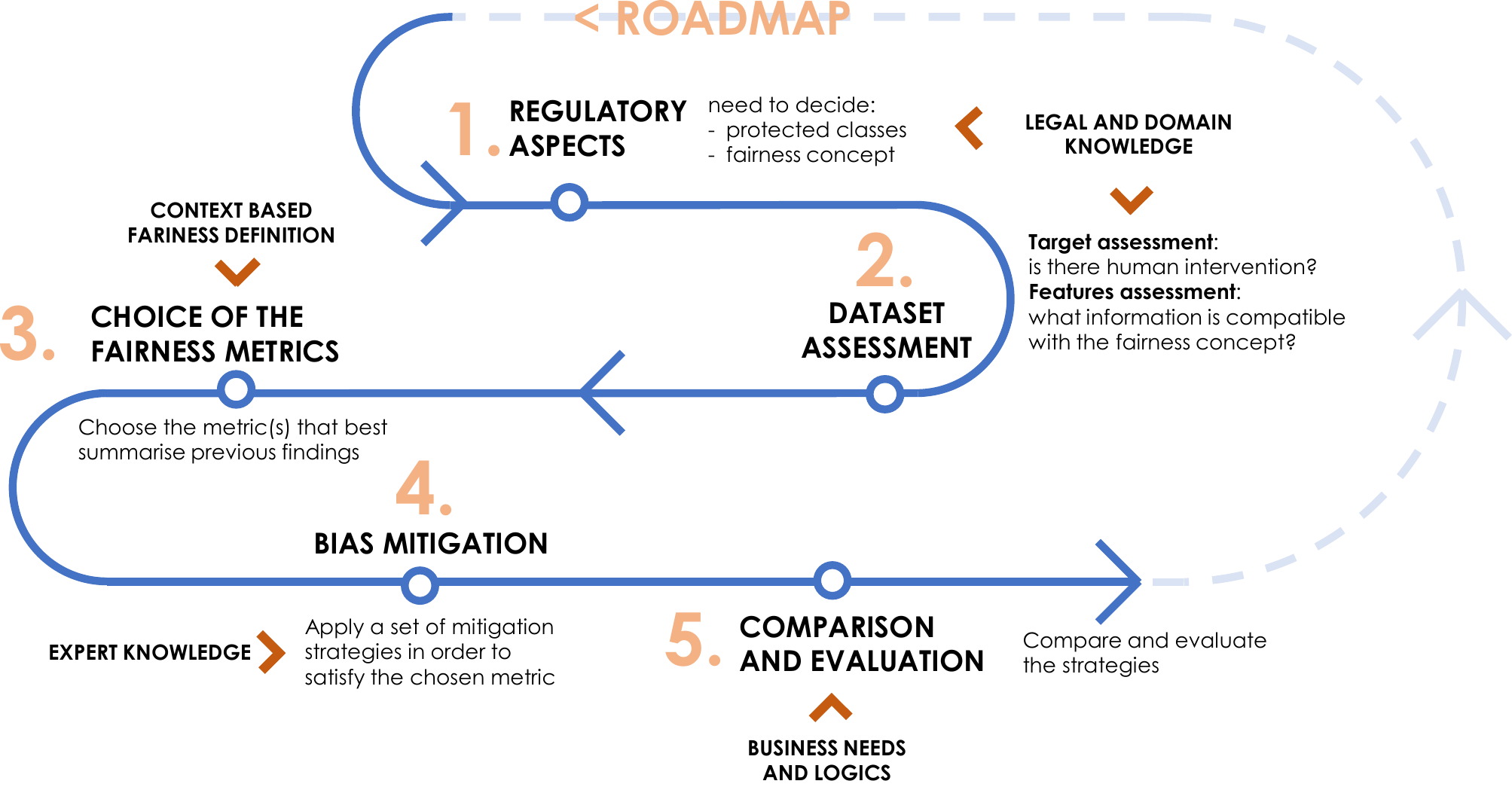}
    \caption{Schematic visualization of the proposed roadmap representing the process of pursuing fairness in ML projects.}
    \label{fig:roadmap}
\end{figure*}

In this section, we present our proposal for a generic roadmap to enable fairness in ML. It encompasses five states: regulatory aspects, dataset assessment, choice of fairness metrics, bias mitigation and comparison/evaluations.

It is notable that the roadmap requires specific inputs from various expertise (such as legal and domain knowledge, expert knowledge, etc.). Indeed, pursuing fairness is a process far too intertwined with several ethical and social aspects to be treated as a purely technical issue.

As a final remark, this roadmap must be thought of as a flexible guideline, with steps back and iterations over specific points, in order to converge to the best possible solution between regulatory aspects, mathematical formulation, algorithmic performance and fairness optimization. The process is at least as important as metric optimization.

The roadmap is outlined in the following steps (see \figurename~\ref{fig:roadmap}):

\subsection{Regulatory aspects} 

[Domain knowledge, legal expertise] As we have seen in section~\ref{sec:background} there is no single notion of fairness and its definition is highly dependent on specific aspects of the use case at hand and of its domain. In order to decide what is the potentially \emph{sensitive information} and the \emph{concept of fairness} to be pursued, it is necessary to take into account legal and regulatory aspects.

However, the fact that, sooner or later, there will be regulations to clearly prescribe what is and what is not fair in each situation is unlikely, unreasonable and in many ways undesirable. Thus, it is crucial that in each domain and even use case, people working on it, being them developers, scientists or domain experts, consider carefully the consequences of including or not potentially sensitive attributes and the variables correlated to them and what could mean, from the point of view of the final user, to be unfairly discriminated, of course taking into account all the regulatory aspects relevant to the specific situation. 
In any case, the advent of regulations is itself a (slow) process, during which companies and service providers can play a role by actively doing research and building their own policies and best practices.

\subsection{Dataset assessment} [Domain knowledge, legal expertise] Once the concept of fairness and the sensitive variables have been defined, it is needed to assess what information in the dataset can be actually used to produce fair decisions. This can be done by trying to answer the following questions: 
\begin{itemize}
    \item \emph{Target variable}: is the ground truth variable the result of some form of human judgement or is it based only on facts? This information is important to understand whether the target variable can or cannot be used to actually measure fairness. Metrics such as Equality of Odds heavily rely on the ground truth target to quantify fairness and cannot be used if the target itself is prone to some form of bias.
        
    Notice that, in the credit lending example, even a variable that may seem objective, as the actual repayment or not of a loan, is subject to a form of selection bias, since it is an information available only for people that were granted a loan in the first place and that are customers of a single bank \cite{yeom2018discriminative}.
        
    \item \emph{Feature variables}: what information in the variables is compatible with the chosen concept of fairness? In the credit lending use case, e.g., it may be that a sensitive attribute like gender or citizenship is correlated with income, but it may be compatible with the chosen concept of fairness to use it nonetheless to make lending decisions, since it may be considered a fair way in which sensitive information impacts the final decision: income is for sure a crucial variable to determine the probability of repayment.

    On the other hand, one may consider that the differences in income (e.g. with respect to gender) are themselves due to historical bias and thus opt for the use of techniques to remove the dependence between income and the sensitive attributes, in order to mitigate this historical bias as well (see e.g. \cite{yeom2018discriminative} and references therein for more insights).
    
    Notice that, if the target variable is deemed to be objective, one may stick to the Equality of Odds fairness metric and in this case the information in the dataset that is compatible with the chosen fairness concept is precisely the one justified by the target itself.
    
    This decision is of course closely interdependent with step 1, namely with the choice of the proper concept of fairness and of the sensitive attributes.
    
\end{itemize}
    
\subsection{Choice of the fairness metric(s)} [Domain knowledge, legal expertise, data science expertise] Given the outcome of steps 1 and 2, it is possible to choose, among available fairness metrics, the one(s) that best embodies the chosen fairness concept, given the dataset assessment. In particular, some choices to be made are:
\begin{itemize}
    \item target dependent / target independent;
    \item group / individual;
    \item observational / causal.
\end{itemize}
Depending on the use case, one may decide to monitor more than one metric, e.g. both a group and an individual notion of fairness. 

In many respects individual and group fairness can be thought of as two extremes of a continuum of possible metrics, roughly depending on what kind of variables one is willing to accept on the basis of the chosen concept of fairness. Namely, one could condition over all the non-sensitive variables, thus enforcing a form of Conditional Demographic Parity which is equivalent to simply removing the sensitive feature only; or one could not condition at all, thus enforcing the group notion of Demographic Parity, e.g. by removing all the information of the sensitive feature present in the dataset (i.e. the variable itself and all its correlations with other variables). Intermediate forms of Conditional Demographic Parity lie between these two extremes.
    
The choice of the metric is a crucial step that summarises the knowledge and expertise of steps 1 and 2 and allows to translate them into a mathematical and algorithmic framework.

\subsection{Bias mitigation} [Data science expertise] Given a specific fairness metric, different strategies can be implemented in order to fit a model by pursuing both algorithmic performance and an optimal value of the chosen metric. As discussed in \ref{sec:background}, these strategies are usually classified in pre-processing, in-processing, post-processing techniques, depending on the specific point of the algorithmic pipeline in which fairness optimization is implemented.
    
Notice that most of the literature on mitigation techniques is focused on group notions of fairness, while only a small fraction is devoted to mitigation of individual forms of bias. Moreover, individual fairness is still a somewhat ambiguous concept, being it usually defined loosely as ``similar people are given similar decisions'', which can be interpreted in many different ways. For instance, the Fairness Through Unawareness is itself a possible form of individual fairness, since two individuals identical with respect to all features but different in the sensitive one(s) are given identical outcomes by design.       

On the other hand, many have criticised group fairness, since it may happen that, in order to reach the desired metric value, two individuals similar in all features but in the sensitive one, are given different outcome. Namely, protected individuals are effectively favoured with respect to other individuals. Thus, the demand for applications of individual fairness is constantly increasing, but, up to now, it has to face the problem of the difficulty of providing a clear and precise way to measure it.
    
\subsection{Comparison of the strategies and evaluation} [Domain knowledge, data science expertise] Once a set of mitigation strategies has been implemented, an evaluation in terms of both algorithmic performance and fairness metric must be made in order to eventually choose the strategy to put in place.

\section{BeFair\label{sec:befair}}

As pointed out in section~\ref{sec:roadmap}, there is no such thing as the fairness metric and thus there is neither a single or definitive way to mitigate bias in all situations.

In order to support and guide data scientists through the steps toward the pursue of fairness, we have developed \texttt{BeFair}: a collection of tools that allows to implement the notional steps presented in the roadmap on a real use case.
The toolkit includes functionalities to detect bias in the input dataset and in model outcomes, to mitigate bias using different strategies derived from the literature, to evaluate the performance of models according to the most relevant metrics, to compare different strategies with a performance/fairness trade-off rationale and to interpret specific feature relationships through a causal graph.

Our work is focused on technical aspects and does not explicitly incorporate the legal expertise prescribed in the roadmap.

In the reminder of this section we delve into the credit lending use case and subsequently introduce the toolkit and its application.

\subsection{Data assessment on credit lending}

Decisions concerning credit lending are indeed highly susceptible to unfairness. To evaluate each application, financial institutions often request a certain amount of personal information whose examination might potentially lead to, even unintentional, discrimination. The dataset we used for implementing \texttt{BeFair} comes from an anonymised portion of past loan granting applications, whose actual final outcome had no dependence in any way on a ML model.

The dataset consists of about $10^5$ loan applications accompanied with a set of personal and financial information, including sensitive attributes, as well as information related to the application, such as the requested amount and duration. Throughout the following sections, we perform the assessment and apply the mitigation strategies over the feature \textit{citizenship} as a mere example, to be able to point out that without intention a standard model can inject discrimination in the predictions even if no significant bias exists in input data.

While there is no specific numerical formula laid out by anti-discrimination laws, to quantitatively determine bias in data we leverage an instantiation of the U.S. Equal Employment Opportunity Commission (EEOC)  \cite{feldman2015certifying}. In the document, they adopt the so-called 80\% rule stating that the ratio between the percentage of subjects belonging to a certain protected group assigned the positive decision outcome and the percentage of subjects not belonging to that group also assigned the positive outcome should be no less than 80:100.  In our running example the data satisfies the requirement since the value of the \textit{disparate impact} measured on the attribute \textit{citizenship} over the original target remains below the value 0.8, or equivalently Demographic Parity below 0.2.

\subsection{Fairness mitigation techniques in credit lending}
\label{sec:befair_mitigation}
For the credit lending use case, we applied a set of fairness mitigation strategies that we briefly describe in the rest of the section and whose results are summarised in Table~\ref{tab:results} and discussed in section~\ref{sec:discussion}.

\subsubsection{Pre-processing}

We implement 3 different pre-processing techniques: suppression, massaging and sampling.
In \emph{Suppression} \cite{kamiran2012data} the transformed dataset is derived by removing both the sensitive variable and the features with highest correlation with it. We removed all variables with correlation higher than 15\%. \emph{Massaging} \cite{kamiran2009classifying, kamiran2012data} consists in relabelling the target label of some observations in order to reach Demographic Parity for the ``massaged'' target. To choose which observations must be relabelled, an auxiliary classifier is trained (the literature proposes to use a Bayesian classifier but we have obtained better results using a Random Forest). Finally, \emph{Sampling} \cite{kamiran2012data} simply consists in over - or under - sampling observations in order to reach Demographic Parity.

Once the dataset has been transformed, we apply a standard Random Forest to get the mitigated outcomes.

Notice that these three techniques are all aimed at reaching Demographic Parity.

\subsubsection{In-processing}

We implement two different algorithms: Adversarial Debiasing and Reductions.

\emph{Adversarial Debiasing} \cite{zhang2018mitigating} is based on the simultaneous training of two competing classifiers (corresponding to a Generative Adversarial Network). 

In the first one, the predictor $P$ tries to accomplish the task of predicting the target variable $Y$ given the input variables $X$ by modifying its weights $W$ to minimise some loss function $L_P(\hat{Y}, Y)$. The second one, the adversary $A$, tries to accomplish the task of predicting the sensitive variable, given $\hat{Y}$ by modifying its weights $U$ to minimise some loss function $L_A(\hat{Z}, Z)$ and consequently backpropagates the error through the predictor $P$.
    
If the Adversary model is trying to estimate the sensitive attribute given only $\hat{Y}$ the result will satisfy Demographic Parity. Rather, if it is trying to estimate the sensitive attribute given $\hat{Y}$ and the true label $Y$, the result will satisfy Equality of Odds.

We rely on the Python module \texttt{AIF360} \cite{aif360-oct-2018} to apply Adversarial Debiasing in order to either impose Demographic Parity (AdvDP), Equality of Odds (AdvEO) or Conditional Demographic Parity (AdvCDP). The latest is achieved by training a different Adversarial model for each subgroup of the variable we condition on. In our use case, we condition on three different levels credit risk: high/medium/low.

\emph{Reductions} approach \cite{agarwal2018reductions} is based on the idea of finding a classifier that minimises the classification error subject to a specific fairness constraint, effectively reducing the problem to a sequence of cost-sensitive classification problems.
More specifically, it takes an arbitrary ML model and trains it multiple times updating, at each iteration, the weights to be assigned to each observations in order to fulfill the chosen fairness constraint. 
It is actually an hybrid between an in-processing method, since it works by imposing fairness during training and a post-processing method, since it can be applied to any ML models, treating them as black-boxes.

We rely on \texttt{Fairlearn} Python module \cite{bird2020fairlearn} for the actual implementation of the Reductions approach. 

We train different models (starting from a logistic regression) using Demographic Parity metric. In particular, first, we train 50 models using the GridSearch algorithm (ReductionsGS) and select the best one. Second, we train 2 models using the ExponentiatedGradient (ReductionsEG) with 0.001 and 0.01 values for the constraints \cite{agarwal2018reductions}.

\subsubsection{Post-processing}

As mentioned in section~\ref{sec:background}, these are a set of techniques that basically consists in computing the mitigated decisions as a function
$\hat{Y} = f(R, A)$, where $R$ is the outcome of any given (in general biased) classifier (in our case a Random Forest) and $A$ the sensitive attribute, such that it satisfies a desired fairness metric.

We implement a simple algorithm to impose Demographic Parity (ThreshDP) and one to impose Equality of Opportunity (ThreshEopp) and we rely on \texttt{Fairlearn} Python module \cite{bird2020fairlearn} to compute post-processing mitigation enforcing Equality of Odds (ThreshEO)\footnote{Demographic Parity and Equality of Opportunity can be easily enforced by computing group-wise thresholds; while, in general, Equality of Odds requires some form of randomization, since it may not be possible for each group to select a singe threshold classifier reaching the same true positive rate and false positive rate. See~\cite{hardt2016equality} and \texttt{Fairlean} documentation~\cite{bird2020fairlearn} for more details.}.
Conditional Demographic Parity (ThreshCDP) is enforced by choosing a threshold not only group-dependent but also depending on the level of the variable we are conditioning on, that is chosen to be a 3-level credit risk, as for the Adversarial Debiasing case.


\subsubsection{Counterfactual fairness through causality}

\texttt{BeFair} includes the implementation and particularization of counterfactual fairness \cite{kusner2017counterfactual} based on the domain knowledge of our specific use case.

Counterfactual fairness requires to establish a causal graph that represents how the variables influences each other (including input features and outcome). This is usually represented by a Directed Acyclic Graph (DAG) in which each variable is represented by a node and arrows represent causal relationships.

\begin{figure}[htbp]
    \centering
    \includegraphics[width=\columnwidth]{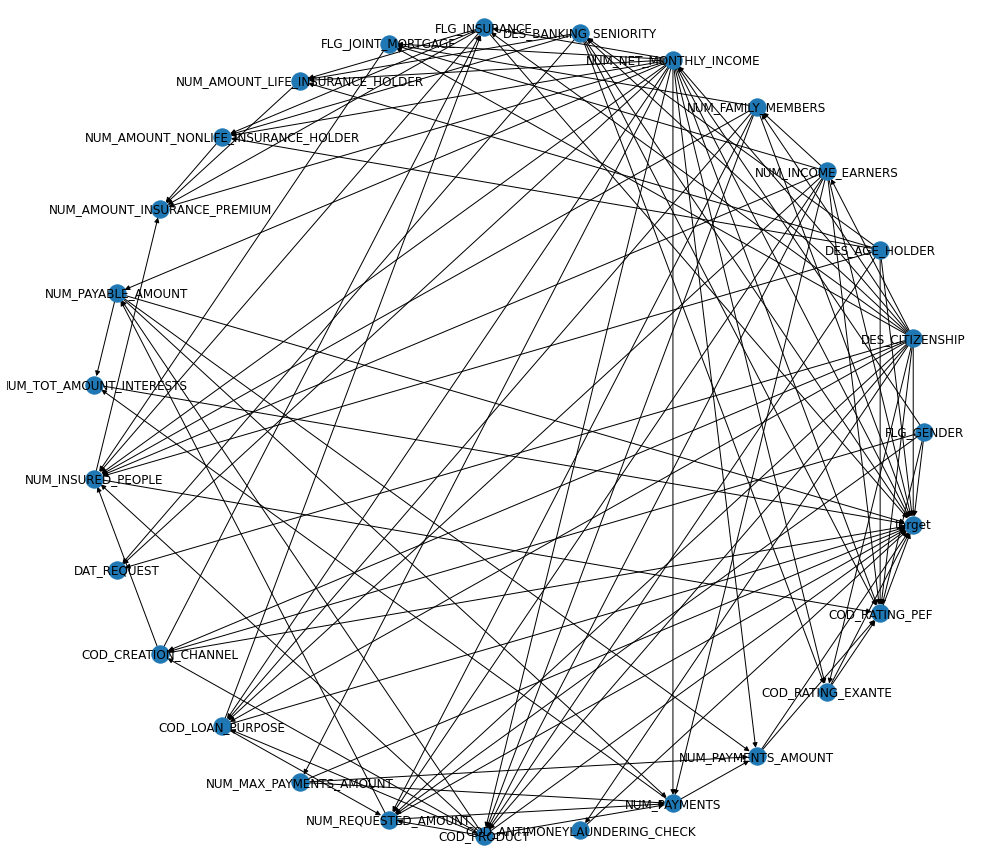}
    \caption{Causal graph for the credit lending use case.}
    \label{fig:causal}
\end{figure}

We have defined our specific causal graph using multiple causal discovery algorithms. In particular, we employed the Python \texttt{Causal Discovery ToolBox}~\cite{kalainathan2019causal}, including different graph modelling algorithms on observational data (i.e. SAM, PC) and the NOTEARS algorithm \cite{zheng2018dags} included in the Python library \texttt{CausalNex}. Once the results were obtained and integrated, a manual revision has been performed to verify the validity of each relation detected (represented by arrows). A group of domain experts participated in the validation process to determine the final causal graph that is included in \texttt{BeFair} (\figurename~\ref{fig:causal}).

Given the defined causal graph, we have developed a Counterfactually Fair model (CFF) based on level 3 described in \cite{kusner2017counterfactual}: the main idea is to model the data using an additive error model with deterministic residuals (error terms of the model) and then fit the CFF-fair model on non-descendants of sensitive attribute(s) and the residuals.
The process to obtain counterfactual outcomes entails to change the value of the sensitive attribute and to propagate the effect of the change throughout the causal graph to obtain the new values of the new input features.
The model trained on the non-descendants of the sensitive feature(s) and the residuals is then counterfactually fair by design, meaning that an individual and its counterfactual version are given the same outcome.

\input{results.tex}

\subsection{Model comparison in credit lending}
\label{sec:optimal_selection}
According to what mentioned in sections \ref{sec:roadmap} and \ref{sec:befair_mitigation} about the vast possibility of mitigation techniques and metrics, we propose two approaches that synthesise the trade-off between them and that might help to identify the most suitable model according to the specific domain:

\begin{itemize}
    \item \emph{Trade-off fairness-performance}. This indicator is inspired by $F_\beta$-score used in ML, whose value for $\beta = 1$ results in the harmonic mean of precision and recall:
    \[
        (1 + \beta^2)\frac{(1 -|\phi|) \times \pi}{\beta^2 \times (1 -|\phi|) + \pi};
    \]
    where $\pi$ and $\phi$ are the preferred performance and fairness metrics, respectively and $\beta$ is the weight associated with the performance metric.

    \item \emph{Constrained performance}. Once chosen an upper bound $\Phi$ for a desired fairness metric $\phi$, this indicator corresponds to the highest possible performance given that fairness constraint 
    \[ 
        \max_{\phi \leq \Phi} \pi.
    \]
\end{itemize}
The optimal choice is then given by the model maximizing the selected indicator.

In our \texttt{BeFair} implementation we have considered the following parameters to compare models, although other metrics could be included easily. For the performance metric $\pi$ we have considered accuracy, precision, recall and F1. For the fairness metric $\phi$ we have included Demographic Parity, Equal Opportunity, Predictive Parity and Equality of Odds.

\begin{figure*}[htbp]
    \centering
    \includegraphics[width=.9\textwidth]{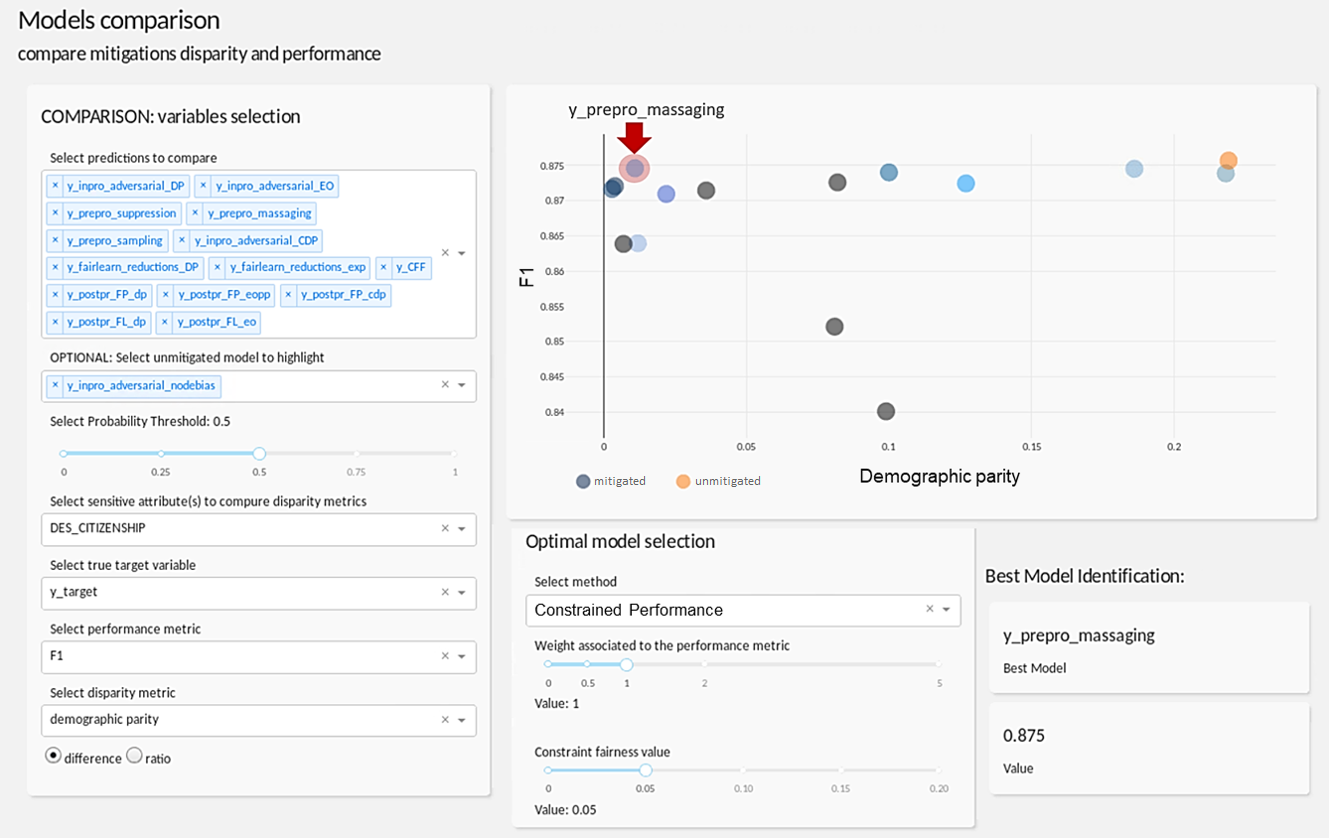}
    \caption{Models comparison: snapshot of \texttt{BeFair}. Several mitigation strategies can be selected, together with the desired protected attribute and the chosen performance and fairness metrics to be used (left panel). The right panel is devoted to the performance-fairness plane (top) and to the selection of the optimal model (bottom) (see~\ref{sec:optimal_selection}).}
    \label{fig:comparison}
\end{figure*}

\texttt{BeFair} allows users to configure the model comparison. \figurename~\ref{fig:comparison} depicts an example of a graph generated by \texttt{BeFair} showing the trade-off among fairness ($x$-axis) and performance ($y$-axis). In particular, it shows the comparison using DP and F1 metrics for all the mitigation strategies implemented (blue dots in the graph) and models without mitigation (orange dots). The best strategy ($y\_prepro\_massaging$) is identified, in this example, using the constrained performance approach with $\Phi = 0.05$, reaching an F1 = 0.875.

\section{Discussion of results\label{sec:discussion}}



Table~\ref{tab:results} summarises the results of the mitigation strategies introduced in section~\ref{sec:befair} relative to the sensitive attribute of \textit{citizenship} for the credit lending use case. 

Fairness quantification is done via the most used statistical fairness dimensions, namely $Y$-independent (DP), recall based (EO, EOpp) and precision based (PP); while performance is monitored via usual metrics such as Area Under the ROC (AUROC), accuracy and the F1 score (i.e. harmonic mean of precision and recall).

First of all, it is clear from the top panel that naively applying a common ML model to the entire dataset results in amplification of bias with respect to almost all the possible fairness dimensions.
This simple fact is \emph{per se} a sufficient reason to promote and foster the need of attention on fairness issues in ML models, in particular in its banking sector applications.

The intuitive and simplest pre-processing methodology of suppression seems not to be able to reach the same level of mitigation of the other strategies, moreover paying a higher price in terms of predictive performance. Thus, bias mitigation while preserving performance is not as simple as removing a bunch of variables from the training dataset.
Better results come from massaging the dataset.

Fairness Through Unawareness (FTU)\footnote{The underlying classifier is a plain Random Forest.}, as expected, is not able to mitigate bias with respect to group metrics, since it does not take into account the sensitive information still present in the dataset via correlations to other variables.

The in-processing techniques are able to reduce DP and EO maintaining the same performance as the unmitigated models. The downside in using these methods is that they consist in training models specifically designed to meet some fairness constraint and they cannot be generalized to arbitrary models. 

Post-processing techniques, selecting group-wise thresholds over a trained model, have binary 0/1 outcomes, thus for them it is not possible to compute the Area under the ROC. However, they seem to be overall pretty good in all fairness dimensions without loosing much in performance. On the other hand, they can be criticised for the fact that they, almost by design, consist in treating protected groups differently.

Further clarification must be made for techniques aimed at imposing Conditional Demographic Parity (AdvCDP, ThreshCDP). These techniques result in rather poor performance when assessed with respect to the fairness dimensions present in Table~\ref{tab:results}.
This is expected: imposing CDP means to go in the direction of a more individual notion of fairness, where equality is requested only for people having some common conditions (in this example the same level of credit risk). This, of course, does not mean that these strategies produce less fair outcomes, it only means that they are fair with respect to a different notion of fairness. This is in line with the discussions in section~\ref{sec:roadmap} about the paramount importance to carefully consider the notion of fairness appropriate for a specific domain and task. As mentioned above, this decision cannot be left entirely to data scientists and developers, since it involves both domain and legal expertise in complex ways. It is highly desirable that people with different expertise work together along the roadmap, especially in its first steps and their iterations, where crucial decisions are taken about the notion of fairness and the information that can be used safely.

A similar argument holds for the Counterfactual model (CFF), which indeed performs poorly in terms of group fairness dimensions, being it a method trying to enforce an individual notion of fairness. 
Moreover, it must be taken into account that counterfactual models are unfalsifiable by design, namely there are many alternative counterfactual realities compatible with the same causal graph and implying in general a different definition of what is fair, and no observation can be used to choose among them. 

However, the process of building and validating a causal graph for a specific use case marks an important step in the comprehension of the network of interdependence among the variables involved. The more these connections and dependencies are known, the more it is possible to understand what it means, in that specific situation, to be unfairly discriminated. Thus, we believe it represents a valuable tool in the roadmap to fairness.

Notice that the fairness dimension related to equal precision (PP) seems to be worsened by most mitigation techniques: this is coherent with the fact that most techniques aim at reducing DP or EO, thus ``loosing ground'' in a measure like PP which is related to precision. Once again, it is worth mentioning that it is not possible to be fair with respect to every possible dimension. 

Even if the performance and fairness requirements for the specific case are well-defined, still it is needed to choose the model that best meets them simultaneously. For this purpose, we have proposed two different approaches: (i) use the \emph{constrained performance} method when the fairness constraint is clearly defined as in case of the EEOC instances, otherwise (ii) use the \emph{trade-off fairness-performance} method.

In general, Table~\ref{tab:results} seems to suggest that in this credit-lending use case it is possible to enforce fairness along different dimensions almost without any deterioration of performance.

\section{Conclusion\label{sec:conclusion}}

Fairness has many dimensions. We have seen that several techniques are available both to assess and to mitigate them. However, it is still not clear which dimension should be pursued in each specific situation.

Our roadmap is an attempt to provide a general guideline to address fairness in ML projects and focuses on the fact that different expertise should work together along the process in order to properly take into account the context and the social impact of the technological service/product being developed.
It is worth highlighting that our roadmap to fairness is independent of the credit lending dataset used for application in this paper, instead it has been conceived to be use case agnostic and generalised.

Our \texttt{BeFair} toolkit allows data scientists and developers to embed several bias mitigation techniques and assessment metrics within their ML projects and to compare these with a rationale of fairness/performance trade-offs. These can be used to eventually take the practical decisions with respect to the aforementioned roadmap.

We showed the use of \texttt{BeFair} to assess bias in real data from a credit lending use case and to compare different mitigation techniques. The results confirm that using ML models without taking measures to avoid unfair outcomes may lead to strong bias amplification and also that each mitigation methodology has its own strengths and limitations and the choice among them is strictly dependent on the fairness dimension one focuses on.

We believe that more research is needed on the ethical and legal side to disentangle and explicitly elaborate on various categorisations of bias that concur to form the overall discrimination in specific domains. It will facilitate in understanding the information that can be safely exploited in different situations.

On the technical side, effort is still needed to understand more clearly the relationship and trade-offs among different fairness metrics, in particular with respect to the group vs. individual dimension. Indeed, despite many attempts, a precise mathematical formulation of individual fairness remains an open challenge and will likely be subject of future research.

\section*{Disclaimer\label{sec:disclaimer}}
The views and opinions expressed within this paper are those of the authors and do not necessarily reflect the official policy or position of Fujitsu Laboratories of Europe and Intesa Sanpaolo. Assumptions made in the analysis, assessments, methodologies, models and results are not reflective of the position of any entity other than the authors.

Furthermore, at the moment of writing, Intesa Sanpaolo does not have any ML algorithm in place to evaluate credit lending applications.

\bibliography{IEEEabrv, ref.bib}

\end{document}

%% file: results.tex
\begin{table*}[!t]
    \ra{1.3}
    \caption{Fairness and performance assessment of mitigation strategies discussed in section~\ref{sec:befair} with respect to different metrics. Each panel is devoted to a particular family of mitigation techniques. Fairness metrics are expressed in terms of difference over sensitive groups, thus the lower in absolute value the better. Bold (underline) highlights the best (worst) value per column. Results are discussed in section~\ref{sec:discussion}.}
    \label{tab:results}
    \centering
    \begin{tabular}{llccccccccc}
        \toprule
        
        & & \phantom{abc} & \multicolumn{4}{c}{\bf fairness} & \phantom{abc} & \multicolumn{3}{c}{\bf performance}\\
        
        \cmidrule{4-7} \cmidrule{9-11}\\[-3ex]
    
        {\bf family} & {\bf type} && {\bf DP} & {\bf EO} & {\bf EOpp} & {\bf PP} && {\bf AUROC} & {\bf Accuracy} & {\bf F1}\\
        \midrule
        
        \multirow{3}{*}{no mitigation} & Logistic && \underline{0.324} & \underline{0.272} & \underline{0.272} & {\bf 0.032} && 0.817 & 0.761 & 0.823 \\
        & Random forest && 0.221 & 0.202 & -0.104 & 0.068 && {\bf 0.838} & 0.804 & 0.875\\
        & Neural network && 0.219 & 0.198 & 0.104 & 0.072 && 0.830 & 0.811 & {\bf 0.876}\\  
        \midrule
        \multirow{3}{*}{pre-process} & FTU && 0.164 & 0.124 & 0.058 & 0.095 && {\bf 0.838} & 0.812 & {\bf 0.876} \\
        & Suppression && 0.099 & -0.053 & 0.065 & 0.152 && \underline{0.753} & \underline{0.748} & 0.840 \\
        & Massaging && -0.004 & 0.062 & 0.062 & 0.163 && 0.818 & {\bf 0.868} & \underline{0.803}\\
        & Sampling && 0.080 & 0.012 & 0.012 & 0.115 && 0.835 & 0.791 & 0.851\\
        &  CFF && 0.218 & 0.192 & 0.104 & 0.070 && 0.832 & 0.810 & 0.874\\
        \midrule
        \multirow{3}{*}{in-process} & AdvDP && -0.034 & 0.073 & 0.063 & \underline{0.176} && 0.823 & 0.802 & 0.869\\
        & AdvEO && 0.102 & 0.029 & -0.010 & 0.148 && 0.819 & 0.805 & 0.871\\
        & AdvCDP && 0.147 & 0.101 & -0.050 & 0.112 && 0.830 & 0.807 & 0.872\\
        & ReductionsGS && 0.012 & 0.077 & 0.049 & 0.159 && 0.812 & 0.794 & 0.864\\
        & ReductionsEG && 0.007 & 0.084 & 0.051 & 0.161 && -- & 0.794 & 0.864\\
        \midrule
        \multirow{3}{*}{post-process} & ThreshDP && {\bf 0.003} & 0.099 & 0.056 & 0.164 && -- & 0.805 & 0.872\\
        & ThreshEO && 0.082 & {\bf 0.006} & 0.006 & 0.138 && -- & 0.812 & 0.873\\
        & ThreshEOpp && 0.100 & 0.048 & {\bf 0.005} & 0.119 && -- & 0.809 & 0.874\\
        & ThreshCDP && 0.186 & 0.159 & 0.072 & 0.083 && -- & 0.810 & 0.875\\
        
        \bottomrule
    \end{tabular}
\end{table*}